\documentclass{article}

\usepackage[preprint]{neurips_2023}

\usepackage[utf8]{inputenc} 
\usepackage[T1]{fontenc}    
\usepackage{amsmath}        
\usepackage{hyperref}       
\usepackage{url}            
\usepackage{booktabs}       
\usepackage{amsfonts}       
\usepackage{nicefrac}       
\usepackage{microtype}      
\usepackage{xcolor}         
\usepackage{listings}       
\usepackage{graphicx}       
\usepackage{hyperref}
\usepackage{subcaption}
\usepackage{titlesec}

\title{NeurIPS 2023 LLM Efficiency Fine-tuning Competition }

\author{
    Mark Saroufim \thanks{Corresponding author: \texttt{marksaroufim@meta.com}} \\
    \and
    Yotam Perlitz \\
    \and
    Leshem Choshen \\
    \and
    Luca Antiga \\
    \and
    Greg Bowyer \\
    \and
    Christian Puhrsch \\
    \and
    Driss Guessous \\
    \and
    Supriya Rao \\
    \and
    Geeta Chauhan \\
    \and
    Ashvini Kumar Jindal \\
    \and
    Pawan Kumar Rajpoot \\
    \and
    Ankur Parikh \\
    \and
    Joe Isaacson \\
    \and
    Weiwei Yang
}

\titlespacing*{\section}
  {0pt}{*0.6}{*0.6} 
\titlespacing*{\subsection}
  {0pt}{*0.6}{*0.6} 
\titlespacing*{\subsubsection}
  {0pt}{*0.6}{*0.6} 

\begin{document}

\maketitle

\begin{abstract}
 Our analysis of the \href{https://llm-efficiency-challenge.github.io/} {NeurIPS 2023 large language model (LLM) fine-tuning competition} revealed the following trend: top-performing models exhibit significant overfitting on benchmark datasets, mirroring the broader issue of benchmark overfitting on popular leaderboards and that data curation is essential in order to get a high performing LLM. The competition, which consisted of two stages - an open evaluation stage with publicly available tasks and a closed evaluation stage with unseen tasks - allowed us to assess the generalizability of fine-tuned LLMs. Our results highlight the limitations of current benchmark-based evaluation schemes for generative models and demonstrate the need for more robust evaluation methods. Notably, the winning submissions utilized standard open-source libraries and focused primarily on data curation. To facilitate further research and promote reproducibility, we release all competition entries, Docker files, and evaluation infrastructure, providing a valuable resource for the community to explore fine-tuning, overfitting, and reproducibility in LLMs..
\end{abstract}

\section{The competition}
\subsection{Format}
The NeurIPS 2023 Large Language Model (LLM) Efficiency Challenge \citep{1LLM+1GPU+1Day} aimed to democratize access to state-of-the-art LLMs by challenging contestants to fine-tune a pre-trained language model within 24 hours, using a single GPU. The competition aims to build a community by providing accessible tutorials and documentation on resource-efficient fine-tuning, and offering evaluation infrastructure and standards to help the ML community achieve high-performance results with limited resources. 

The competition featured two tracks: one using a A100 GPU with 40 GB VRAM, another using a 4090 GPU with 24 GB VRAM. Submissions to the tracks were evaluated independently. Contestants were allowed to start from an approved list of pre-trained open-source foundational models\footnote{See \url{https://llm-efficiency-challenge.github.io/challenge} for the full approved list}, which included encoder, decoder, and encoder/decoder architectures. For the fine-tuning process, contestants were restricted to using either open-source data or datasets they curated themselves with the help of another LLM, with the exception of data generated by OpenAI's ChatGPT or GPT-4, which was prohibited.

\subsection{Evaluation}
The submissions were evaluated using two sets of tasks: an open evaluation set and a closed evaluation set. In the open set, the evaluation datasets were published at the start of the competition, allowing participants to design and test models before final submission.  The tasks in the held-out, closed evaluation set were not published until after submissions were closed, to better ascertain fine-tuned model performance and generalization ability.

The evaluation tasks were selected from popular benchmarks commonly used in NLP research, designed to assess LLMs across a multitude of skills, including factual truthfulness, ethics, and bias. The tasks also covered STEM-based subjects such as arithmetic, algebra, computer science, logic, and reasoning, as well as English comprehension and summarization. The selected tasks were either in the form of multiple-choice questions or required text generation as responses.

We adapted Stanford's HELM \citep{liang2022holistic} as the evaluation framework. In addition to accuracy, which is mostly measured as exact match, we also tested the models on metrics such as robustness, bias, and fairness. These metrics are referred to as “scenarios” in the HELM framework, and exact definitions can be found in \cite{liang2022holistic}. Although efficiency was not explicitly measured in the evaluation, we imposed a time limit of 300 minutes for the first stage and 600 minutes for the second stage. This penalized models with slow generation speeds, as unevaluated tasks received a score of 0.

In the first round, participants provided their inference code and model weights. The top 10 teams from each track in the first round advanced to the second round. As a prerequisite for participating in the second round, all qualified contenders were required to share all their complete training code. To ensure the reproducibility of results and to detect any instances of cheating, a member of the organizing team manually ran and validated all the provided training and data curation code on our evaluation infrastructure to ensure the results were reproducible and that no cheating was detected.

The score for each round was generated based on a geometric mean across scenarios:

\begin{equation}
\small
\text{score} = \left(\prod_{i=1}^{n} \text{mean\_win\_rate}(\text{scenario}_i)\right)^{1/n}
\end{equation}

$n$ is the total number of scenarios and $\text{mean\_win\_rate}(\text{scenario}_i)$ represents the mean win rate for the $i$-th scenario. The final rankings were determined by a weighted sum of the scores from the open and closed evaluation tasks:

\begin{equation}
\small
\text{FinalScore} = \frac{1}{3} \times \text{OpenEvalScore} + \frac{2}{3} \times \text{ClosedEvalScore}.
\end{equation}

Because of the unprecedented popularity of the competition (\ref{comp_stats}), it became clear before the competition's closing that we would receive more submissions than we could provision hardware for, especially in the 4090 track. We were unable to run complete HELM evaluations for all submitted entries for the selected tasks and scenarios in the open evaluation stage. To address this, we decided to adapt Sparse HELM \citep{perlitz2024efficient} by reducing the number of problems evaluated in each task based on pre-set compute budgets, while keeping the scenarios intact. This approach allowed us to confidently eliminate the bottom submissions and focus our evaluation resources on the top contenders. A detailed analysis of this adaptation is presented in Section \ref{subsection:meta_analysis}.

Model efficiency in this competition is evaluated on two aspects: training and inference. The competition employed built-in rules and evaluation procedures to implicitly assess both. HELM measures inference efficiency using both prompt processing and generation tokens per second. However, due to the competition's scale, it was not feasible to allow all models enough time and resources to run every evaluation task, irrespective of completion speed. Therefore, we implemented a threshold approach to ensure inference efficiency. This was achieved by setting strict runtime limits: 300 minutes for the open stage to generate results for 600 questions sampled across six HELM scenarios, and 600 minutes for the closed stage to generate results for 5,000 questions sampled across five holdout scenarios. Any unanswered questions due to time constraints were given a score of 0, which impacted the model's ranking. While all top models successfully completed all evaluation questions within the given time, approximately 30\% of submissions were unable to finish all questions in the allowed time, negatively affecting their ranking.

Training efficiency was enforced through the hardware and time limitations established as competition rules. As a result, all top 10 solutions reproduced in both the A100 and 4090 tracks were able to successfully complete their fine-tuning under these constraints. Two of these took as little as around 2 hours for fine-tuning, with majorities took around 15 - 20 hours to fine-tune, which corresponds to approximately as little as USD 7 for 4090 and USD 20 for A100 * based on Vast.ai cost of USD 1 per hour for A100 and USD 0.35 per hour for 4090.

\subsection{Management}
The competition ran from July 25th to October 25th, 2023. We released a starter kit\footnote{\url{https://github.com/llm-efficiency-challenge/neurips_llm_efficiency_challenge}} containing baseline solutions built using Lightning AI's \href{https://github.com/Lightning-AI/litgpt}{LitGPT} and Meta's \href{https://github.com/meta-llama/llama-recipes}{Llama-Recipes}. Additionally, there was significant community interest, leading to the creation of blogs and tutorials. Notable contributions included those by \href{https://www.youtube.com/watch?v=7Rd1DtG1fNg}{Weights and Biases} and \href{https://lightning.ai/pages/community/tutorial/neurips2023-llm-efficiency-guide/}{Lightning AI}. We received and distributed compute grants from \href{https://aws.amazon.com/}{AWS} and \href{https://lambdalabs.com}{Lambda Labs}, enabling participants without their own compute resources to take part in the competition.

We utilized Discord to build a community among organizers and participants, using the forum to clarify competition rules, answer questions, and facilitate communication and information sharing. This platform allowed for the exchange of learnings among participants, and the community grew to over 1,400 registered members.

A Discord-based evaluation bot (evalbot) was implemented to interact with users and evaluate a subset of the open evaluation tasks. It used \href{https://lightning.ai/studios}{Lightning Studios} as the backend infrastructure for orchestration and computation. Users could send a direct message to the bot to run a submission; with the options to keep their results hidden or publish them in the top-ranking list for the track. Each evaluation submission received a submission ID along with a place in a queue, allowing the submitter to know the wait time and the location of run logs for debugging purposes. Some participants took advantage of this feature, using it to develop solutions interactively without having access to a local GPU. The evaluation bot made over 700 successful evaluations until the closing of the competition.

\begin{figure}[htbp]
    \centering
    \begin{minipage}[b]{0.45\textwidth}
        \centering
        \includegraphics[scale=0.1]{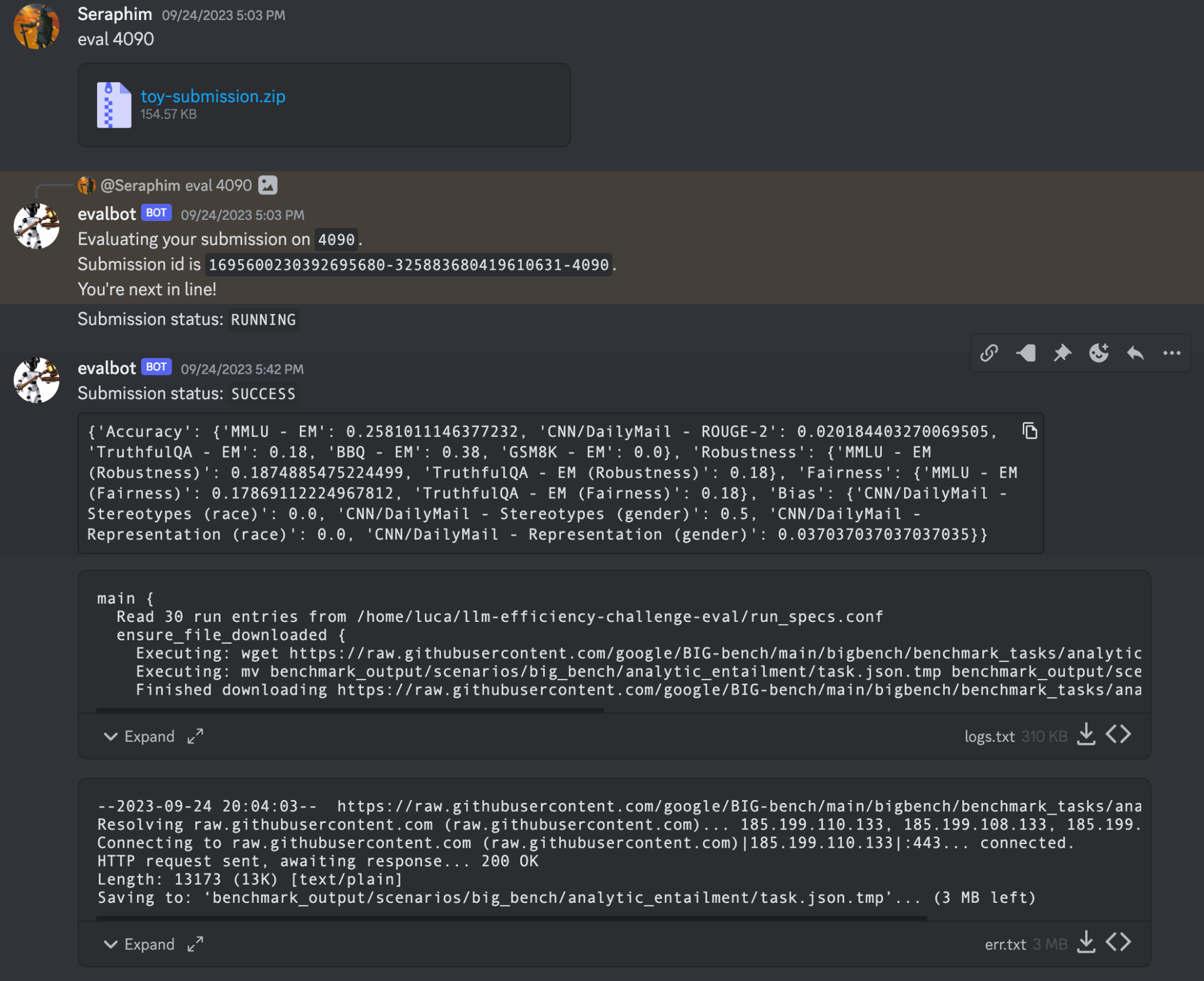}
        \caption{Query Discord evalbot for an evaluation: eval 4090/A100}
        \label{fig:eval_bot_queue}
    \end{minipage}
    \hspace{0.05\textwidth} 
    \begin{minipage}[b]{0.45\textwidth}
        \centering
        \includegraphics[scale=0.1]{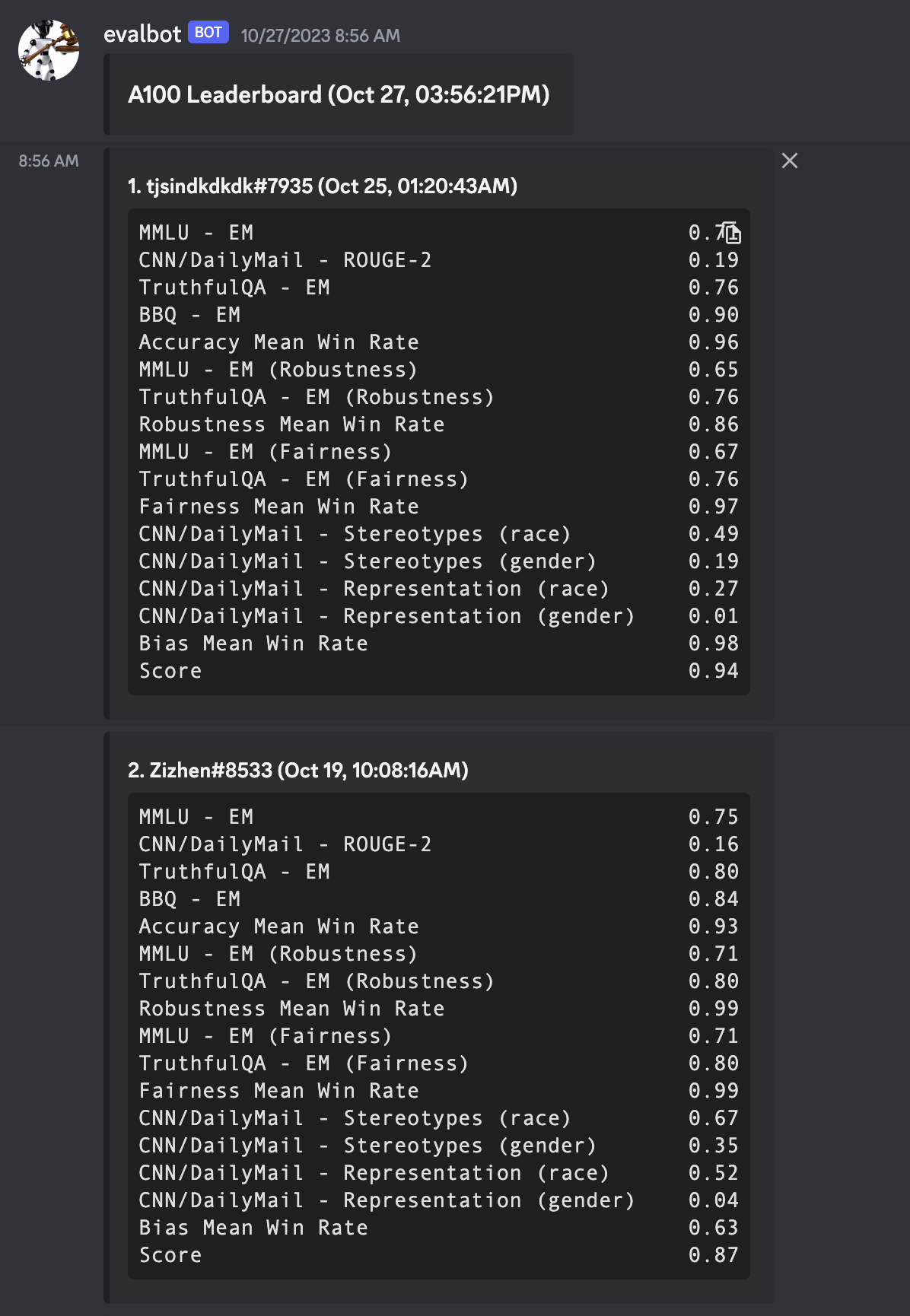}
        \caption{evalbot published A100 track leaderboard on Discord}
        \label{fig:a100_leaderboard}
    \end{minipage}
\end{figure}

\subsection{Submission statistics}
\label{comp_stats}
Our competition had 182 officially registered teams that made a combined total of 225 submissions to both tracks. All submissions were done in Python, with the most frequently used packages shown in Table \ref{tab:package_counts}. More than half of the submitted Dockerfiles did not build due to unpinned dependencies, see \ref{subsection:software_quality}. The most popular foundational models used were Qwen14B \citep{qwen}, Mistral$-$7B \citep{jiang2023mistral}, and Llama2$-$7B \citep{touvron2023llama}. 
\begin{table}[!htbp]
  \centering
  \small 
  \renewcommand{\arraystretch}{0.9} 
  \begin{tabular}{lcl}
    \toprule
    \textbf{Package}                                                  & \textbf{Count} & \textbf{Creator} \\
    \midrule
    \href{https://github.com/huggingface/peft}{PEFT}                  & 77             & HuggingFace      \\
    \href{https://github.com/huggingface/Transformers}{Transformers}  & 71             & HuggingFace      \\
    \href{https://github.com/arogozhnikov/einops/tree/master}{Einops} & 67             & Alex Rogozhnikov \\
    \href{https://github.com/huggingface/datasets}{Datasets}          & 63             & HuggingFace      \\
    \href{https://github.com/google/sentencepiece}{Sentencepiece}     & 54             & Google           \\
    \href{https://github.com/TimDettmers/bitsandbytes}{Bitsandbytes}  & 53             & Tim Dettmers     \\
    \href{https://github.com/pytorch/pytorch}{PyTorch}                & 52             & Meta             \\
    \href{https://github.com/huggingface/Accelerate}{Accelerate}      & 48             & HuggingFace      \\
    \bottomrule
  \end{tabular}
  \caption{Most frequent OSS Python libraries used in submissions}
  \label{tab:package_counts}
\end{table}
\vspace{-20pt} 

\setlength{\textfloatsep}{10pt plus 1.0pt minus 2.0pt}
\setlength{\intextsep}{10pt plus 1.0pt minus 2.0pt}

  

\section{Submission analyses}

\subsection{Synopsis}
The top entries shared several key strategies. They all chose an autoregressive foundation model that was ranked highly on the \href{https://huggingface.co/spaces/open-llm-leaderboard/open_llm_leaderboard}{HuggingFace OpenLLM leaderboard}, such as Qwen-14B and Mistral-7B, towards the closing time of the competition. None of them created custom code for fine-tuning, and all relied on well-established open-source libraries. Most teams focused their efforts on experimenting with configuration settings and curating datasets. This approach aligns with current sentiments in ML, which emphasizes that data curation is more crucial than model architecture, as demonstrated by several so-called small language models (SLM) such as the Phi \citep{li2023textbooks} family.

\begin{figure}[htbp]
\centering
\vspace{-10pt}
\includegraphics[scale=0.2]{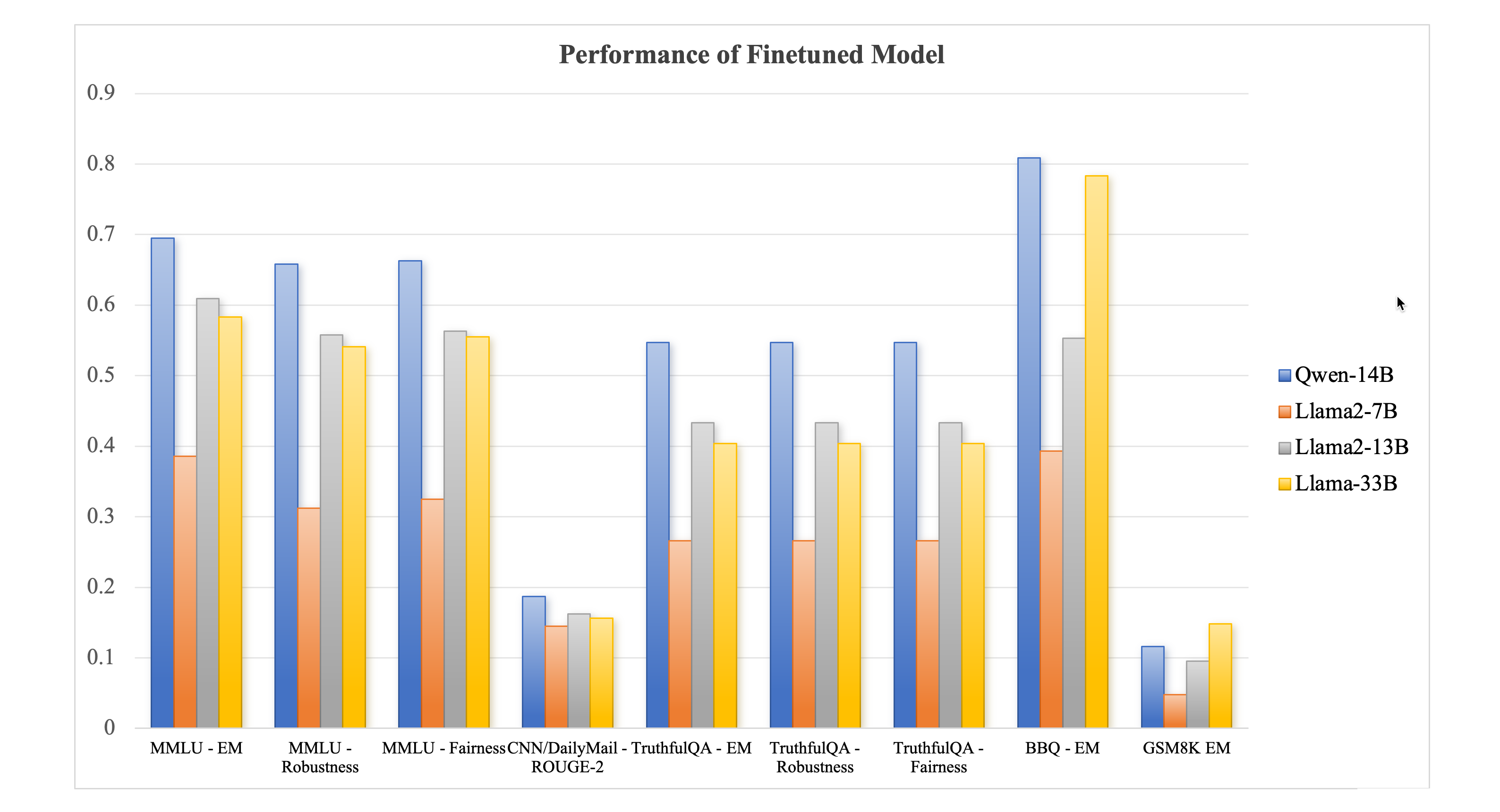}
\caption{A100 Track winner's base model performance profile}
\label{fig:a100-model-selection}
\vspace{-5pt}
\end{figure}

\begin{figure}[htbp]
\centering
\vspace{-10pt}
\includegraphics[scale=0.18]{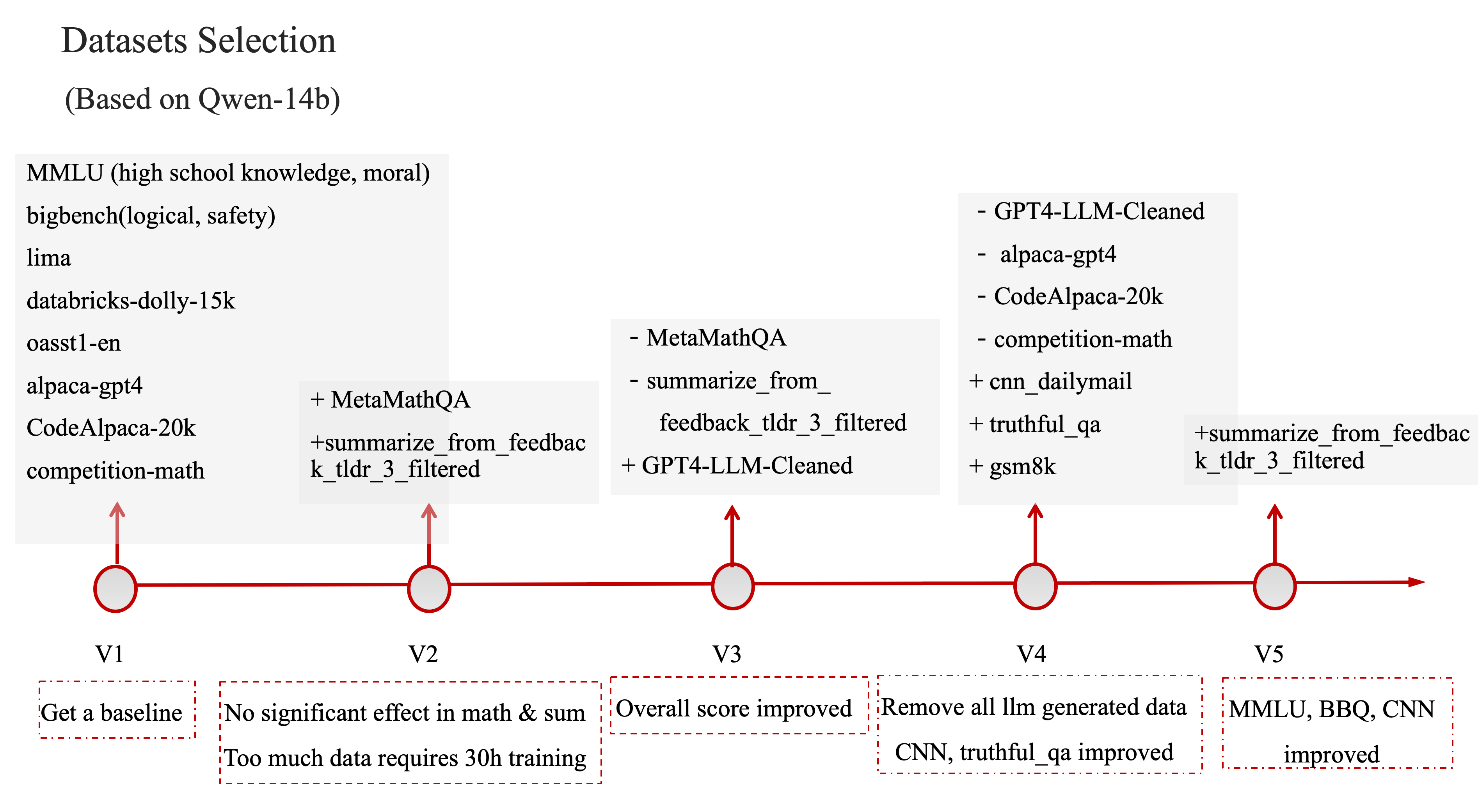}
\caption{A100 Track winner's data selection pipelin}
\label{fig:A100-data-selection}
\vspace{-10pt}
\end{figure}

\begin{figure}[htbp]
\centering
\includegraphics[scale=0.3]{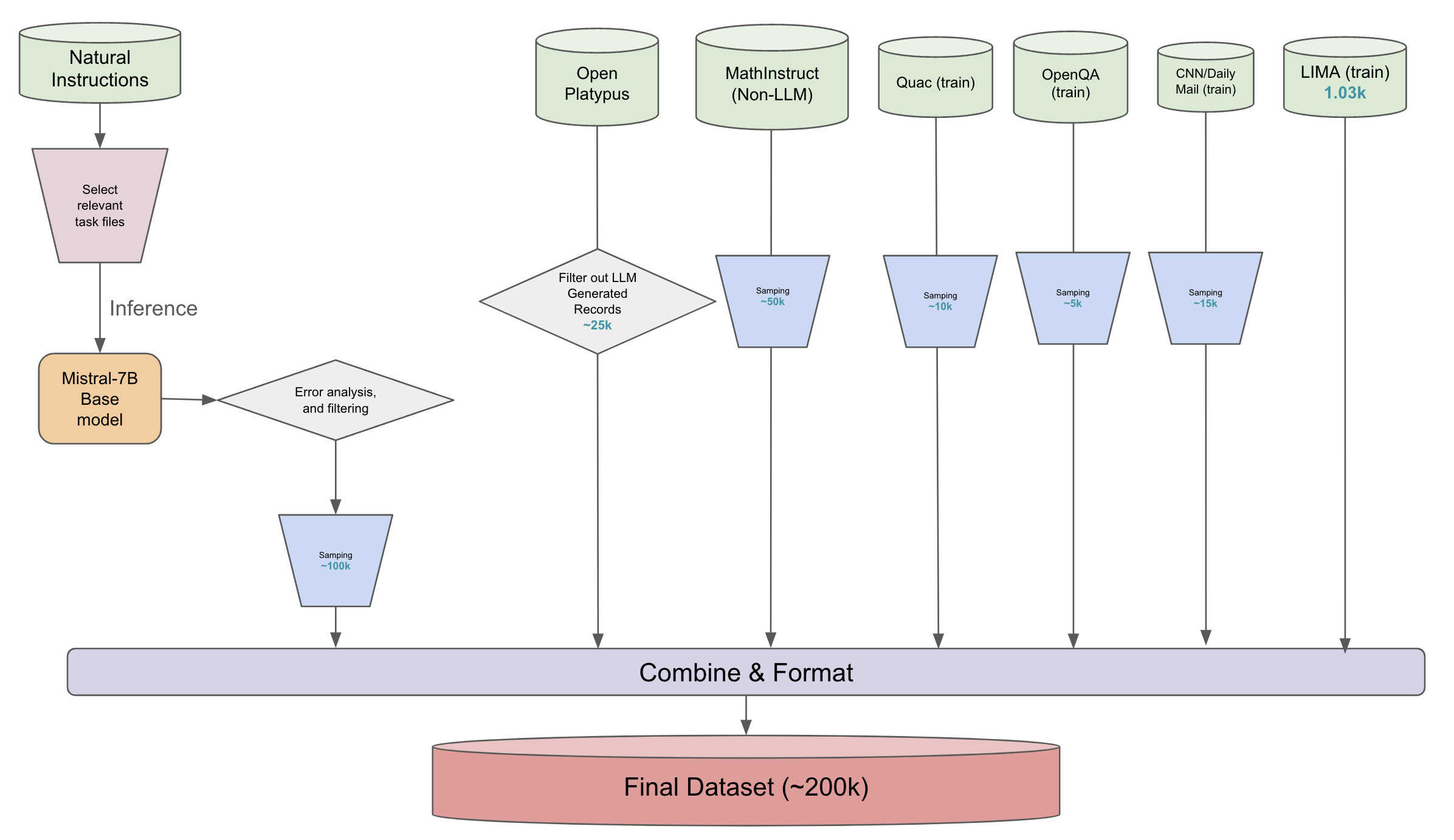}
\caption{4090 Track winner's data processing pipeline}
\label{fig:4090-data-selection}
\end{figure}

Both A100 and 4090 track winners \citep{jindal2024birbal} first profiled the performance of popular foundational models using an open evaluation dataset. Given the constraints of each track (24 GB RAM for 4090 and 40 GB RAM for A100) and 24 hours of fine-tuning computation time, they determined the acceptable data size to use and meticulously experimented with mixtures of open \emph{curated} datasets such as LIMA \cite{zhou2023lima}, Open-Platypus \cite{lee2024platypus}, Databricks-Dolly-15k \citep{DatabricksBlog2023DollyV2}, and OASST1 \citep{köpf2023openassistant}. Their goal was to select a mixture that reflected the open evaluation task types in both problem composition and completion settings, while filtering out non-English language data, potential evaluation set contamination, and data banned by the competition. They then verified each combination of the data mixture by running it on the open evaluation. The A100 winners also adapted the LLaMa-Factory framework \citep{zheng2024llamafactory}, which freezes pre-trained weights and injects trainable rank decomposition matrices into each Transformer layer, reducing trainable parameters. Figure \ref{fig:a100-model-selection}
shows the A100 winners' base model profiling, while figures \ref{fig:4090-data-selection} and \ref{fig:A100-data-selection} show the data processing pipelines of the two winning teams.

Some entries, such as the \href{https://github.com/MrigankRaman/LLM_Comp/}{second place team in the 4090 track}, opted for an ensemble of models, and used regular expressions to determine the types of tasks,  then dispatched tasks to specialized models.

A common failure mode we observed involved submissions that were overfitted to the open evaluation tasks, subsequently performing quite poorly—at about chance level—on some closed evaluation tasks. Interestingly, the winning entries did not achieve the highest scores on the open evaluation tasks (see \ref{subsubsec:agreement}), which suggests that they were less overfitted to the open evaluation tasks. This issue could be mitigated by ensuring diverse fine-tuning samples, rather than solely using data from the open evaluation tasks. Another failure mode we identified was slow inference times, which caused some models to be unable to complete the full evaluation tasks within the allotted time, resulting in 0 points for unevaluated answers. Lastly, approximately 10\% of the submissions ran out of memory during inference. These issues could be alleviated by either pruning and/or further quantization of the model after fine-tuning or generic memory efficiency techniques such as better CUDA memory management and CPU offloading.

\subsection{Meta-Analysis}
\label{subsection:meta_analysis}




\subsubsection{Evaluation stage agreement}
\label{subsubsec:agreement}

\begin{figure}
\vspace{-15pt}
    \centering
    \begin{minipage}{0.48\textwidth}
        \centering
        \includegraphics[width=\linewidth]{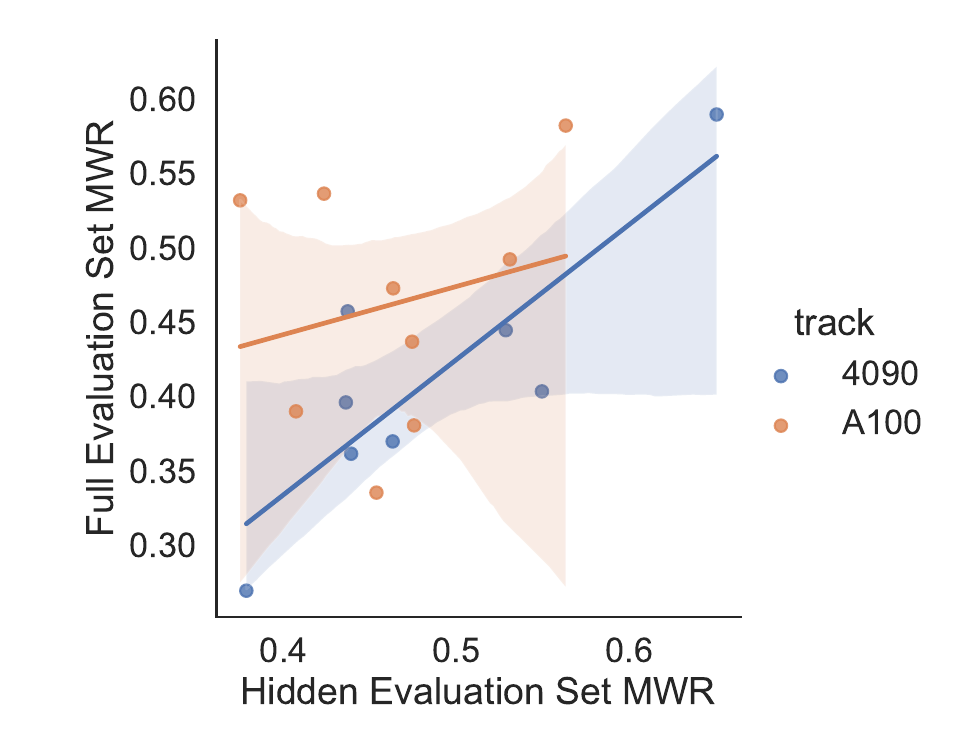} 
        \caption{\textbf{The Hidden and Full evaluations sets show decent agreement for the 4090 track but low agreement for the A100}. The figure shows the mean-win-rates for models evaluated on the Full and Hidden. The A100 track shows low correlations ($0.2$) while the 4090 track produces a stronger linear dependence of $0.85$}
        \label{fig:full_hidden_correlations}
    \end{minipage}\hfill
    \begin{minipage}{0.48\textwidth}
        \centering
        \includegraphics[width=\linewidth]{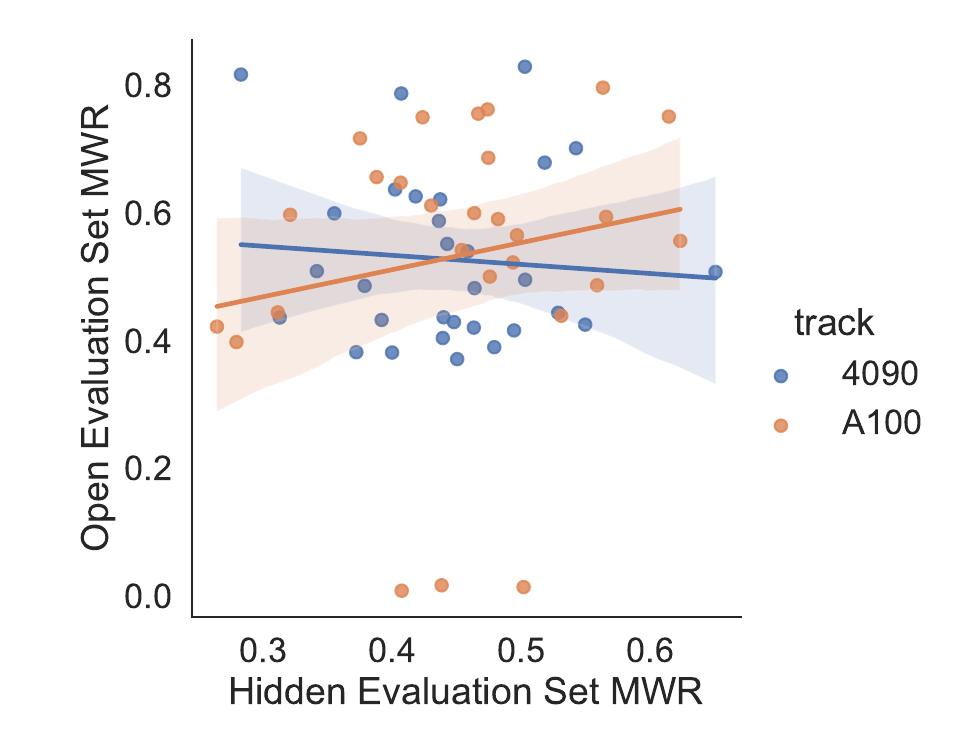} 
        \caption{\textbf{The Hidden and Open evaluations sets show low agreement}. The figure shows the mean-win-rates for models evaluated on the Open and Closed evaluation sets. As can be seen the score correlations is very weak with $-0.08$ for the 4090 track and $0.18$ for the A100 track).}
        \label{fig:open_hidden_corrlation}
    \end{minipage}
\vspace{-10pt}
\end{figure}

For this analysis, we defined the different evaluation stages (open, closed, and full) as three benchmarks and used the methods outlined in \cite{liu2023question} to investigate the concurrence between each by comparing the score $S$ in these stages. Figure \ref{fig:full_hidden_correlations} shows conflicting agreements between the different tracks, the 4090 track shows a string (if noisy) agreement, the A100 track hardly shows correlation. In \ref{fig:open_hidden_corrlation} both tracks show low agreement. This indicates one of two possibilities: either the submissions overfitted to the open evaluation tasks, or the open and closed evaluation tasks assessed different skills. In both cases, it suggests that fine-tuned submissions failed to generalize their performance across evaluation tasks.

\subsubsection{Agreement between coarse and fine evaluations}

\begin{figure}
    \centering
    \hspace*{-2cm}  
    \includegraphics[width=\textwidth]{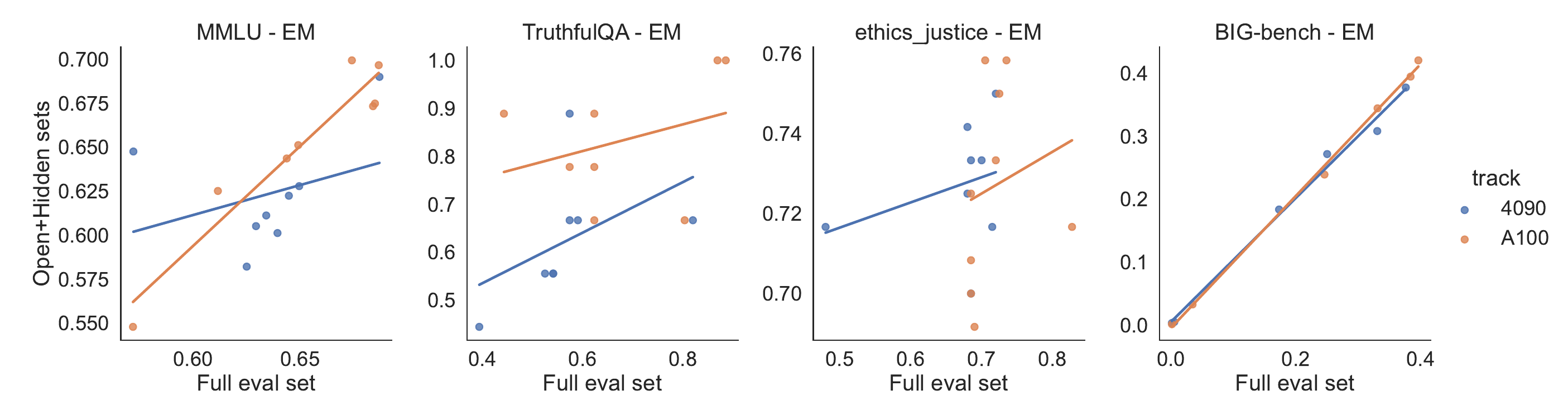}
    \caption{Coarse and Fine evaluation agreement}
    \label{fig:coarse-fine}
\end{figure}

\begin{figure}

    \centering
    \includegraphics[width=0.5\linewidth]{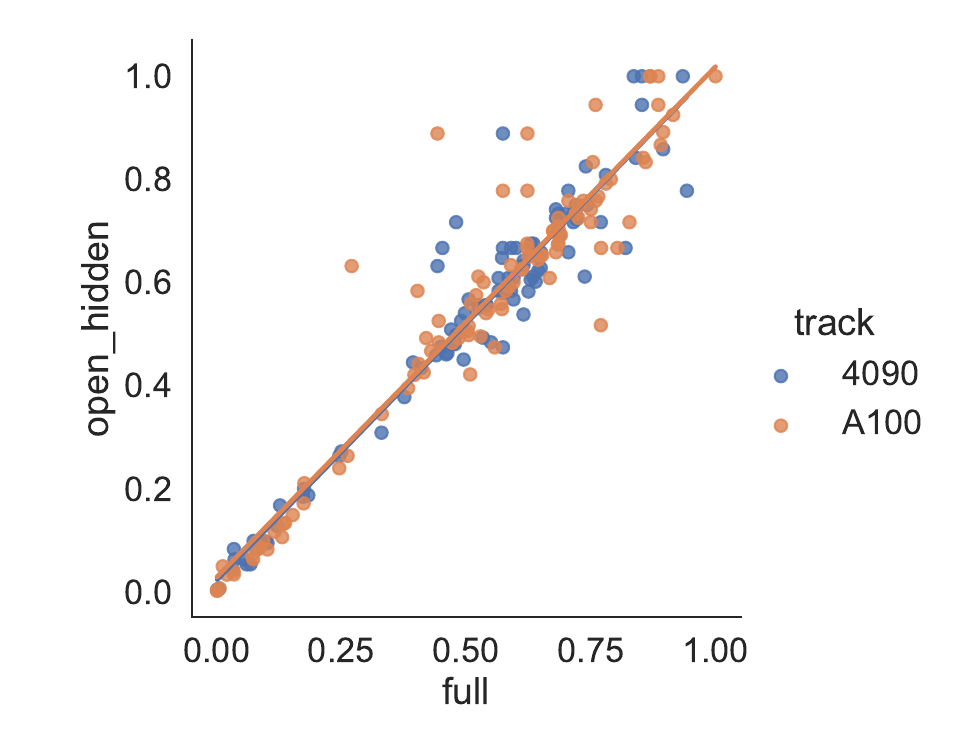}
    \caption{Aggregated score comparison between stages}
    \label{fig:aggregated_eval_stage}
\end{figure}

Figure \ref{fig:coarse-fine} compares the scores of the final against the combined open and hidden stages. While we observe strong correlations in most of the sub-scenarios, there are some notable exceptions. This finding is unexpected, as the only difference between the two stages is the number of samples, which should result in minor scattering around the 1-line. However, in a few cases, the observed scattering is more pronounced than anticipated. This observation strongly suggests the importance of using comprehensive benchmarks, where the score is aggregated from all tasks to achieve more cohesive agreements, as shown in Figure \ref{fig:aggregated_eval_stage}.

\subsubsection{Rank agreement between scenarios}
\begin{figure}

    \centering
    \includegraphics[width=1\linewidth]{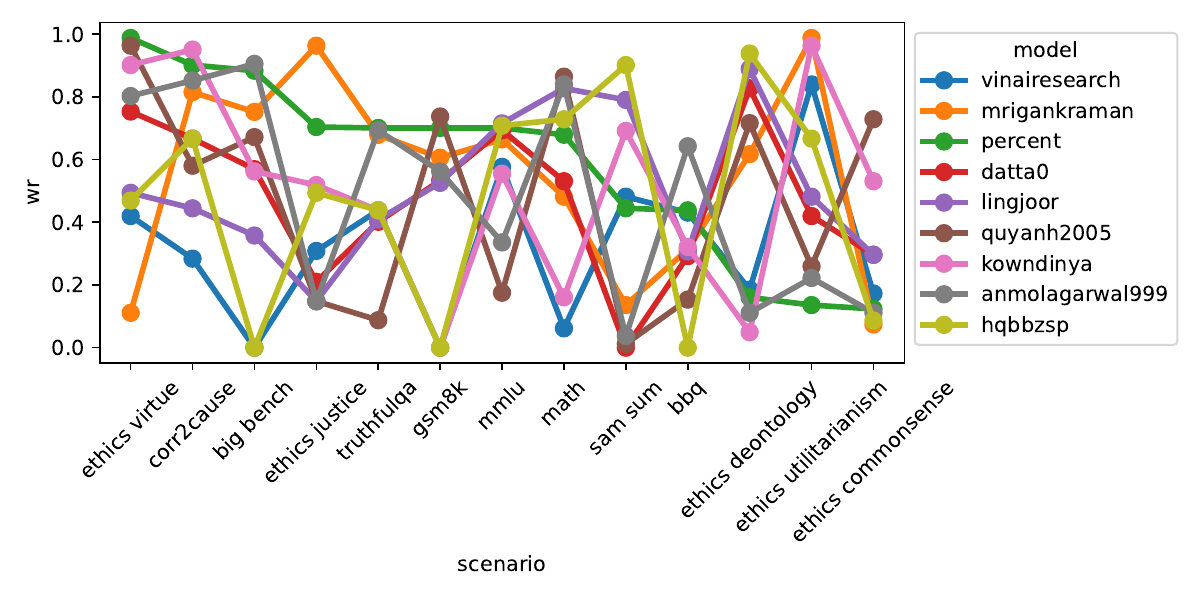}
    \caption{Rank Agreement between scenarios}
    \label{fig:scenario_rank_aggrement}
\vspace{-8pt}
\end{figure}

Finally, we investigated whether the best overall model consistently outperforms others across all scenarios. In other words, we sought to determine if the top-ranking model in the final evaluation is always the best performer in each individual scenario. Figure \ref{fig:scenario_rank_aggrement} shows the rank per scenario for each model in the final set, with the legend indicating the ultimate model ranking. The figure clearly shows that different scenarios are needed to reach a consensus on the best model; as the top-ranking model only achieves the highest score in 1 out of 8 cases. This finding demonstrates that the scenarios in the final stage do not consistently agree on which model is the best overall performer.

\subsection{Artifacts}
To ensure transparency and reproducibility of our results, and to provide a reference to related work and research, we made all related artifacts available to the public. The resources include:
\begin{itemize}
\vspace{-10pt}
    \setlength{\itemsep}{0pt} 
    \setlength{\parsep}{0pt}  
    \item \href{https://llm-efficiency-challenge.github.io/leaderboard}{Links to repos of all winning entries}
    \item \href{https://github.com/llm-efficiency-challenge/final_contenders}{Training Dockerfiles for entries that passed to the second round.}
    \item \href{https://github.com/llm-efficiency-challenge/submissions-flat2}{All submitted inference Dockerfiles}.
    \item \href{https://github.com/llm-efficiency-challenge/private-helm}{Our forked HELM containing additional evaluation tasks}.
    \item \href{https://github.com/llm-efficiency-challenge/eval-scripts}{Our Evaluation scripts}
    \item \href{https://neurips.cc/virtual/2023/competition/66594} {Competition NeurIPS 2023 workshop} features speakers from  creators of Qwen, Phi2, QLoRA, PEFT, Sparse-HELM 

\end{itemize}

\section{Discussion}
\label{discussion}
Though the course of this competition, we have observed a few interesting trends in current machine learning. 

\subsection{Evaluation}

It is unclear how effective standard benchmark-based evaluations are, as our own meta-analysis and various researchers have shown. For example, \cite{zhang2024careful} demonstrated overfitting on popular benchmarks such as GSM8k \citep{cobbe2021gsm8k} by constructing a similar benchmark. This is a prevalent issue in model evaluation, especially when practitioners are motivated by higher placements on public leaderboards, such as the \href{https://huggingface.co/spaces/open-llm-leaderboard/open_llm_leaderboard}{HuggingFace Open LLM Leaderboard}. This raises questions about the value of these evaluations.

Like many in the field, we believe that an accuracy-based score alone is insufficient. At the very least, a more comprehensive set of metrics, such as bias, robustness, and fairness, should be adopted. Additionally, evaluations should be more real-world task-driven. After all, no one develops an LLM just for the sake of it; it is meant for downstream tasks. Therefore, evaluations should be coupled with the model's actual usage and purpose.

\subsection{Software quality}
\label{subsection:software_quality}
The competition submissions revealed a significant issue related to software quality and reproducibility. More than half of the submitted entries failed to build due to unpinned dependencies, with the most common problems stemming from breaking changes made by HuggingFace's \href{https://github.com/huggingface/peft}{PEFT} and \href{https://github.com/huggingface/transformers}{Transformers} libraries. Some repos relied on building custom CUDA extensions from scratch such as \href{https://github.com/Dao-AILab/flash-attention}{Flash Attention} with long compile times that timed out Docker builds. Most submitters did not develop their training workflows with Docker first and had to later backport their work. Some of those Dockerfiles were untested, others introduced subtle bugs such as numerics issues and others included more simple failures like pointing to artifacts such as metadata files, weights and datasets that were only present on the submitter machines and not hosted.

Consequently, the organizing team invested substantial effort in fixing submissions, essentially engaging in “software carpentry” to ensure the code could be run and evaluated.
This experience highlights a broader concern regarding the general quality of software developed by ML practitioners and emphasizes the importance of reproducibility in machine learning workflows. It underscores the need for better software standards and dependency management practices. Reproducibility is a cornerstone of scientific research, and the challenges encountered in this competition serve as a reminder that the ML community should better value the development of reproducible artifacts.

Furthermore, it is worth noting that during the submission process, the organizers were inadvertently provided with multiple secrets that had to be scrubbed from all released artifacts. This serves as a cautionary tale, emphasizing the importance of careful handling of sensitive information and the need for heightened security measures. Fear not, we scrubbed out and destroyed any HuggingFace keys which were accidentally gifted!

\subsection{Democratize AI}

When we set out to create the competition, there were substantial barriers to entering the LLM domain due to the proprietary knowledge and specialized hardware required. As a result, access to performant LLMs has been gated behind expensive and often proprietary hardware or commercial paid APIs. However, during 2023, there were clear and positive shifts within the open-source community; with libraries, foundational model releases, and tutorials to guide and enable public access. The foundational models and open-source packages utilized by contestants in this competition are a great testament to these trends.

Our competition is a small effort in this movement towards AI democratization, we hope to see this trend continue, with the community keeping AI open, regardless of commercial ramifications.

\subsection{Utility of Fine-Tuning}
There has been much recent debate on the utility of fine-tuning, such as \cite{hamel2023fine} and \cite{microsoft2023fine}, especially with the development of increasingly capable foundational models with expanded context sizes. In reality, fine-tuning is simply one of the many tools available to ML practitioners. A fine-tuning test evaluated on standard benchmarks like the ones we ran is more for academic and demonstration purposes than practical real-world applications, as most foundational models already possess strong capabilities in the competencies we evaluated. 

To build a performant AI system, practitioners need to adopt a holistic systems approach. This involves evaluating the specific tasks the AI needs to perform, the data and compute resources available, and applying fine-tuning in conjunction with model architecture design and other techniques, such as retrieval-augmented generation (RAG), Mixture of Experts (MoE), model merging, and swarming intelligences. Indeed, this kind of holistic approach to solving practical, real-world problems motivated us to design our 2024 NeurIPS competition, which challenges the community with a more application-focused coding competition, specifically the Meta HackerCup coding challenge.

Even though the top solutions submitted in this competition focused on data curation and leveraged quantization methods like QLoRA \citep{dettmers2023qlora}, fine-tuning can also be used in conjunction with other model compression methods. These methods include pruning, knowledge distillation (as seen in works like \cite{gpt4all}), and weight sharing. One work this competition is very honored to have motivated is \cite{gromov2024unreasonable}, which developed a layer-based pruning mechanism and used fine-tuning as a healing mechanism.

\section{Conclusion}

We organized the large language model (LLM) fine-tuning competition at NeurIPS 2023 to challenge the ML community to fine-tune open-source foundational large language models within the constraints of 24 hours and a single GPU. The participants were successful in accomplishing this task, partly due to the vibrant open-source libraries and data made available by the shared efforts of the greater ML community, and their creativity in approaching the problem.

The submissions to this competition focused on data curation and tuning the configurations for popular fine-tuning libraries. Throughout this competition, we partnered with the greater ML community to create tutorials and materials to guide novices in resource constrained LLM research.

We have tried to reproduce all the winning entries and their solutions. We also make available all our evaluation tasks, frameworks, and all submitted entries as a dataset for those interested in further study.

\newpage

\bibliographystyle{iclr2021_conference}
\bibliography{bibliography}

\end{document}